\documentclass[a4paper, 10pt, conference]{ieeeconf}      

\IEEEoverridecommandlockouts                              

\usepackage{xcolor}

\usepackage[utf8]{inputenc}

\usepackage{graphicx}
\usepackage{amsmath}
\usepackage{graphicx}
\usepackage{support-caption}
\usepackage{subcaption}
\usepackage{algorithmic}
\DeclareMathOperator*{\argmax}{arg\,max} 
\title{\LARGE \bf
To each route its own ETA: A generative modeling framework for ETA prediction
}

\author{Charul and Pravesh Biyani\thanks{Charul and P. Biyani are with the Department of Electronics and Communication Engineering, Indraprastha Institute of Information Technology, Delhi,
India{\tt\small (e-mails: charuli@iiitd.ac.in, praveshb@iiitd.ac.in)}}}%

\begin{document}

\maketitle
\thispagestyle{empty}
\pagestyle{empty}

\maketitle
\begin{abstract}
	Accurate expected time of arrival (ETA) information is crucial in maintaining the quality of service of public transit.  Recent advances in artificial intelligence (AI) has led to more effective models for ETA estimation that rely heavily on a large GPS datasets. More importantly, these are mainly cabs based datasets which may not be fit for bus based public transport. Consequently, the latest methods may not be applicable for ETA estimation in cities with the absence of large training data data set. On the other hand, the ETA estimation problem in many cities needs to be solved in the absence of big datasets that also contains outliers, anomalies and may be incomplete. This work presents a simple but robust model for ETA estimation for a  bus route that only relies on the historical data of the particular route. We propose a system that generates ETA information for a trip and updates it as the trip progresses based on the real-time information.  We train a deep learning based generative model that learns the probability distribution of ETA data across trips and conditional on the current trip information updates the ETA information on the go.  Our plug and play model not only captures the non-linearity of the task well but that any transit agency can use without needing any other external data source. The experiments run over three routes data collected in the city of Delhi illustrates the promise of our approach.
\end{abstract}

\section{INTRODUCTION}
The quality of any public transport system is crucially dependent on its reliability.  One of the primary metrics that measure the reliability of a public transit is the predictability of its vehicles in reaching specific pre-determined locations (bus stops in case of a bus-based transit).  A predictable and a reliable public transportation also attracts more users \cite{brakewood2018literature} thereby increasing the economic viability of the transit as well as reducing congestion in the road, a huge urban challenge, especially in the developing world.  From the point of view of the transit operators, predictability is key to maintaining its efficiency.  For instance, predicting bus-bunching in real time also helps in avoiding it.  From the passengers perspective, predictability is generally measured by the accurate determination of the expected time of arrival (ETA) of a vehicle on a route. With the advent of smart phones and GPS enabled public transport, it is now easier to show the ETAs of various routes on the phone applications, especially of those routes which are not very frequent.  Not just for the bus based public transport routes, the ETA determination is necessary and vital even for the private on-demand transportation companies like Shuttl (India) and Chariot (US and India), as well as cab companies such as Lyft, Uber, and Ola.  \par 
The ETA estimation problem is as follows. Given the current location of the vehicle and its future trajectory, what is the expected time of its arrival in a given set of locations?  In this paper, we focus the ETA estimation problem only for bus based public transport routes, especially those that make frequent stops. One of the ways to estimate the ETA to a location is to predict the future traffic speed using the historic speed data on the given route. This approach, however, has several limitations.  The first limitation is the lack of availability of the future traffic speed data across the city network. Large commercial entities like Google and Microsoft provide APIs that provide a prediction for future speed data in a city, but they may be both expensive and inaccurate. More importantly, these APIs are heavily biased towards cars.  In reality, the speed profile for buses is often different than that of cars. The other significant limitation in using these speed profiles is the fact that a bus would often make many stops, which needs to be accounted in the calculations for ETA. In summary, the ETA calculation would be erroneous due to the unavailability of stopping time and inaccurate road speed information. In this work, we argue that one can build a simple yet effective system for predicting ETA for a given bus route by using only the past GPS trajectories of a given route. \par 
As noted earlier,  a bus route consists of set of pre-determined locations, also called bus stops. Each route comprises of many trips (up and down) throughout the day.  We assume that all the route vehicles are customised with GPS units. The location information from these GPS units is then used to calculate the time taken to travel between the consecutive stops thereby making a database of historic and actual ETA of the route.  Given this data of historic ETA data set and the current location of the bus in a trip, the problem is to ``predict" in real-time the ETA for all the remaining stops in the trip.  Note that the historic (and ongoing) data of ETA would constitute both travel timings as well as the stoppage timings specific only to the bus route.  \par 
%
%
\par 
In this work, we employ what is known as generative autoregressive models to predict the ETA in a given trip conditional on the travel times between stops so far in the trip. The generative model is trained using the historic data for the previous trips on the same route.   To this end, we form an ETA matrix, whose rows are the trips on a given day and whose columns contain the time taken to travel between consecutive bus stops. As an example, an ETA matrix corresponding to one day for a route operating 20 trips a day with 20 stops will be of the size $20 \times 20$.  We learn the joint distribution of the underlying ETA data using a generative model based on a CNN and predict the ETA in real-time while the route is operational.  Our approach is inspired by pixel-CNN \cite{oord2016pixel} which is used to learn the distribution of natural images (which are also matrices) in order to complete an incomplete image.  However,  unlike images,  the ETA matrix has a temporal (causal) nature and each row of the matrix can only be filled sequentially.  We modify the pixel-CNN approach to suit the special structure in the ETA estimation task. We introduce masking to make our system causal.  Masking also helps in controlling the level of dependencies from past used for the prediction of future values. We explored two different mask instead of using the traditional mask proposed in \cite{oord2016pixel}. One of the contribution of our work is to make the masking operation automatic thereby making the system plug and play. The generative model used in our paper has been demonstrated \cite{oord2016pixel} to explicitly model complex probability distributions that fits the training data and handles both noisy as well as the missing data case well. Through this approach, we integrate both stopping time as well as time travel time in the training data itself. To the best of our knowledge, the transportation researchers have not explored conditional generative models for either traffic prediction or ETA prediction tasks.  The results indicate that we can make a simple and reliable ETA prediction mechanism by using only bus GPS data without relying on any other external speed data. \par 
In summary, the main contributions presented in the work are:
\begin{itemize}
	\item We propose a novel ETA prediction algorithm based on generative autoregressive model that integrates both traffic speed corresponding to the bus as well as the stopping time in one framework. 
	\item The algorithm utilizes only the historic GPS data from the same route and can be independently implemented irrespective of availability of traffic data in the city.  The algorithm is adaptive in nature and can also be implemented in real-time. 
	\item The key algorithm parameters can be tuned automatically thereby increasing the ease of implementation in the real-world scenarios.
	\item Finally, we implement our algorithm on the real-world transit data available in Delhi, India and it outperforms other traffic prediction based ETA estimation algorithms.
\end{itemize}
\section{Related Work}
Most ETA estimation works \cite{sevlian2010travel,de2008traffic,asif2014spatiotemporal,rahmani2013route,billings2006application} in the literature are based on future speed prediction relying on the historic speed data from cars/cabs across the network.  As noted before, the structure of the city network available to the car based traffic speed prediction models may not be valid for the bus routes which also need to allow for stopping times at the bus stops.  For instance, authors in \cite{billings2006application} employ an autoregressive moving average (ARIMA)  model that relies on the fact the future link (road segment) speeds can be accurately predicted by a linear combination of past speeds from few other links. The variability of link speed profiles of buses from our data set indicate a much higher dimension. Moreover, many of the above methods rely on structured ``big" data which simply may not be available for bus route networks.  Further, some previous approaches based on individual road segment based travel time estimation assumes that the travel time on consecutive road segments as independent \cite{westgate2013travel, hofleitner2012learning, sevlian2010travel,pan2012utilizing,de2008traffic, asif2014spatiotemporal, lv2015traffic, rahmani2013route}, which may not always be true as illustrated in recent works based on machine learning. \par 
ML techniques based on $K$ nearest neighbor approach was used to predict the travel time in \cite{rice2001simple,myung2011travel}.  Authors in \cite{chien2003dynamic} suggest a Kalman filtering method. In \cite{wu2003travel}, support vector regression was suggested for predicting the travel time. Gradient boosting method was proposed in \cite{zhang2015gradient}. However, the above methods mainly incorporate temporal nature of the data, while the spatial dependencies were not explicitly modeled.  \par 
Recently, deep learning based methods have been proposed and have shown state of the art performance. Recurrent neural networks (RNN) in \cite{zeng2013development,wang2016traffic} and long short term memory (LSTM) in \cite{{duan2016travel}} have been proposed to predict the travel time. Spatio-Temporal Hidden Markov Models (STHMM) are used to model correlations among different traffic time series in \cite{yang2013travel}.  Recently, deep end to end travel time estimation (DeepTTE) \cite{deepTTE} uses raw GPS and also captures the local spatial dependencies like weather conditions, driving habits, start time, day of the week and an RNN  is used to learn the temporal dependencies on the feature map generated by the geo convolutional layer.  This model predicts the travel time for a complete path, but when generating the individual estimation of road segments, it does not incorporate the spatial correlation in the road segments. All the deep learning model need an extensive data set to train the non-linearity in the models. They also need considerable effort in turning parameters.   \par
Closer to the problem taken in this paper, authors in \cite{lee2012http} and \cite{wang2016simple} use historical bus trajectories for predicting speed across future road segments. 
\section{Methodology}
\subsection{Problem Formulation}
We first discuss the problem formulation for the ETA estimation problem. To this end, we first introduce key notations. A bus route is defined as an ordered list of bus stops $(k=1,...K)$, where $k=1$ is the source and $k=K$ is the destination stop respectively. Each route is undertaken by several trips in a day where each trip ideally should run according to a predetermined timetable. But, these trips may not adhere to the timetable due to various reasons. Let there be a total of $T$ such trips in a route. 
We denote the travel time between stops $k-1$ to $k$ by $t_{\tau,k}$ during the  $\tau^{th}$ trip where, $(\tau=1,\dots, T)$. The ETA estimation problem tackled in this paper is the following. Assuming the $\tau^{th}$ trip is in progress, and the bus is near the $k^{th}$ stop, what is the ETA for the remaining stops on the route.
Mathematically, we would like to predict $\hat{t}_{\tau,(k+\Delta)}$ where $\Delta =\{1,....K-k\}$. We assume the availability of the historical travel time data for the bus route. Note that historical data could be both noisy as well as incomplete. As we shall see later, we use this historical data to train our prediction model. Further, we also assume travel times of the current trip ${t}_{\tau,(1:k-1)}$, and the previous trips {\it of the same day} ${t}_{(1:\tau-1),(1:k-1)}$ is also available. Mathematically, given the prediction model, we use ${t}_{\tau,(1:k-1)}$ and the previous trips i.e. ${t}_{(1:\tau-1),(1:k-1)}$ to predict $\hat{t}_{\tau,(k+\Delta)}$, for $\Delta= 1$ to $K-k$. \par 
One way of estimating the ETA ${t}_{\tau,(1:k)}$ for the $\tau^{th}$ trip is to employ the Maximum a posteriori (MAP) estimation method.
\begin{align} \label{eqn:map}
\hat{t}_{\tau,(k+\Delta)}= \argmax_{\hat{t}_{\tau,(k+\Delta)}} \, p\left(\,\hat{t}_{\tau,(k+\Delta)}\,|\, {t}_{\tau,(1:k-1)},{t}_{(1:\tau-1),(1:k-1)}\,\right).
\end{align}
where, $p(.)$ denote the conditional probability distribution function (pdf) of the travel times from stop $k$ to $k+\Delta$, $\Delta=1$ \text{to} $K-k$, at the $\tau^{th}$ trip conditional on previous travel times of the current trip and the previous trips on the same day. The above method is also optimal assuming the above mentioned conditional pdfs are known and the MAP estimator can be evaluated. However, estimation of such vast number of pdfs for each route is intractable and the MAP computation becomes impractical. To alleviate the above issues, we employ what is known as the generative modeling approach to learn the conditional distributions required in \eqref{eqn:map} in a single and unified framework thereby making ETA estimation sufficiently accurate while enabling low complexity computations that require less data. These learned distributions is then sampled to estimate the ETA values for the future stops in a trip.
%

\subsection{Generative Modeling for traffic prediction}
 Generative modeling is a powerful way of estimating the distribution of the data in an unsupervised way. Last few years have witnessed tremendous research efforts towards generative modeling of the data \cite{doersch2016tutorial, goodfellow2014generative, oord2016pixel}. Generative models generally employ deep learning frameworks to learn an approximation of the true distribution of the data. In this paper, we utilise an autoregressive generative model to explicitly learn the distribution of the ETA data.  \par
\begin{figure}[ht]
	\centering
	\includegraphics[width=0.8\linewidth]{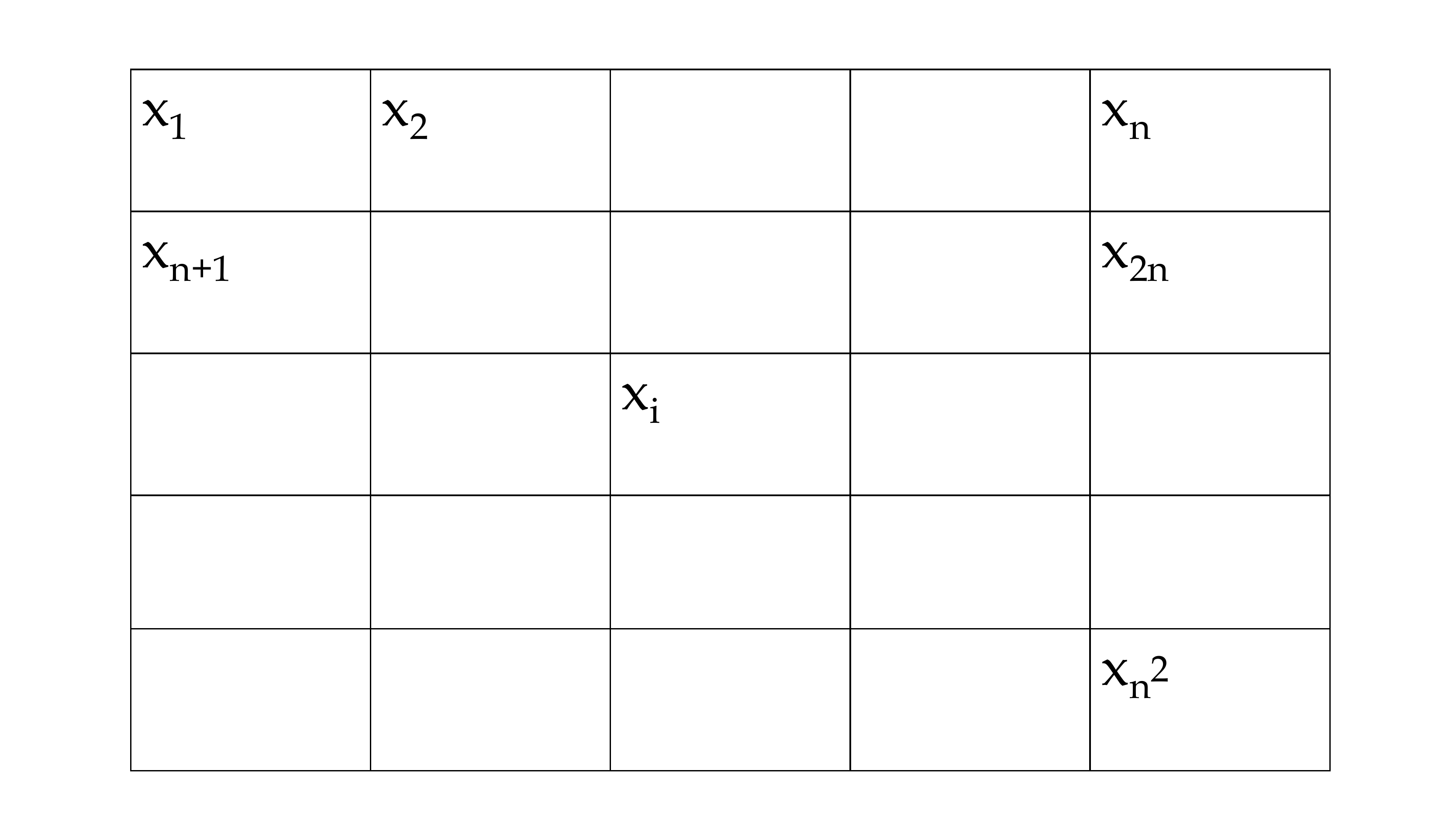}
	\caption{Matrix X }
	\label{fig_img}
\end{figure}

 
Our model is inspired by pixel-CNN \cite{oord2016pixel}, where the authors learn a distribution of natural images. An image is nothing but a matrix of pixel values. The distribution of natural images is thus the joint distribution of all the pixels of a matrix. Formally, let ${\bf X} =(x_1,x_2,x_3, ...,x_{n^2})$ be a matrix shown in Fig. \ref{fig_img}.
Using the chain-rule, the joint distribution of the matrix $\boldsymbol{X}$ is
\begin{align} \label{eqn:probability}
\ p({\bf X})=\prod_{i=1}^{n^2}\, p \left(x_i \,| \,x_{1},....,x_{i-1} \right)
\end{align}
In words, if $\bf X$ is an image, the very first pixel $x_1$ is independent, albeit with a distribution, the second pixel $x_2$ dependent on first, third depends on the first and second and so on. In summary, the matrix is ordered as a sequence of points where the probability distribution of one point depends on the observed values of the previous points. The generation proceeds row by row and pixel by pixel. Similarly, we can determine the probability of pixel $x_i$ conditioned on $x_{i-1}...x_{1}$. \par 
Likewise, the travel times of a bus route can be seen as an image of size $T \times K$ with rows as trips and the columns denote the travel times between consecutive stops. In a way, one day of a bus route travel time matrix can be seen as a single image. Consequently, we can learn the distribution of these ETA matrices using generative models like in \cite{oord2016pixel} by suitably modifying to suit the specifics of the ETA estimation problem. To learn the generative model, we use mask-convolutional neural networks (CNN) based autoregressive model. CNN based models are well known and widely studied for capturing local correlation in images for classification tasks \cite{krizhevsky2012imagenet}.  \par 
The architecture of mask-CNN is fundamentally simple with two blocks in it --- the training and the real-time ETA estimation blocks.  The mask-CNN is utilised to learn the generative model in the training block using the historical route travel time data. Once the model is trained, we use it to compute ETA for a given route. The observed ETA values after the completion of the trip is fed to the training block which simultaneously adapts the model that is used for the future trips. We first briefly describe CNN, followed by detailed discussion of the usage of these models for the ETA estimation problem. 
Before discussing the mask-CNN framework in detail, we provide a simple example to discuss inference based on generative models. 
\begin{figure}[ht]
	\centering
	\includegraphics[width=1\linewidth]{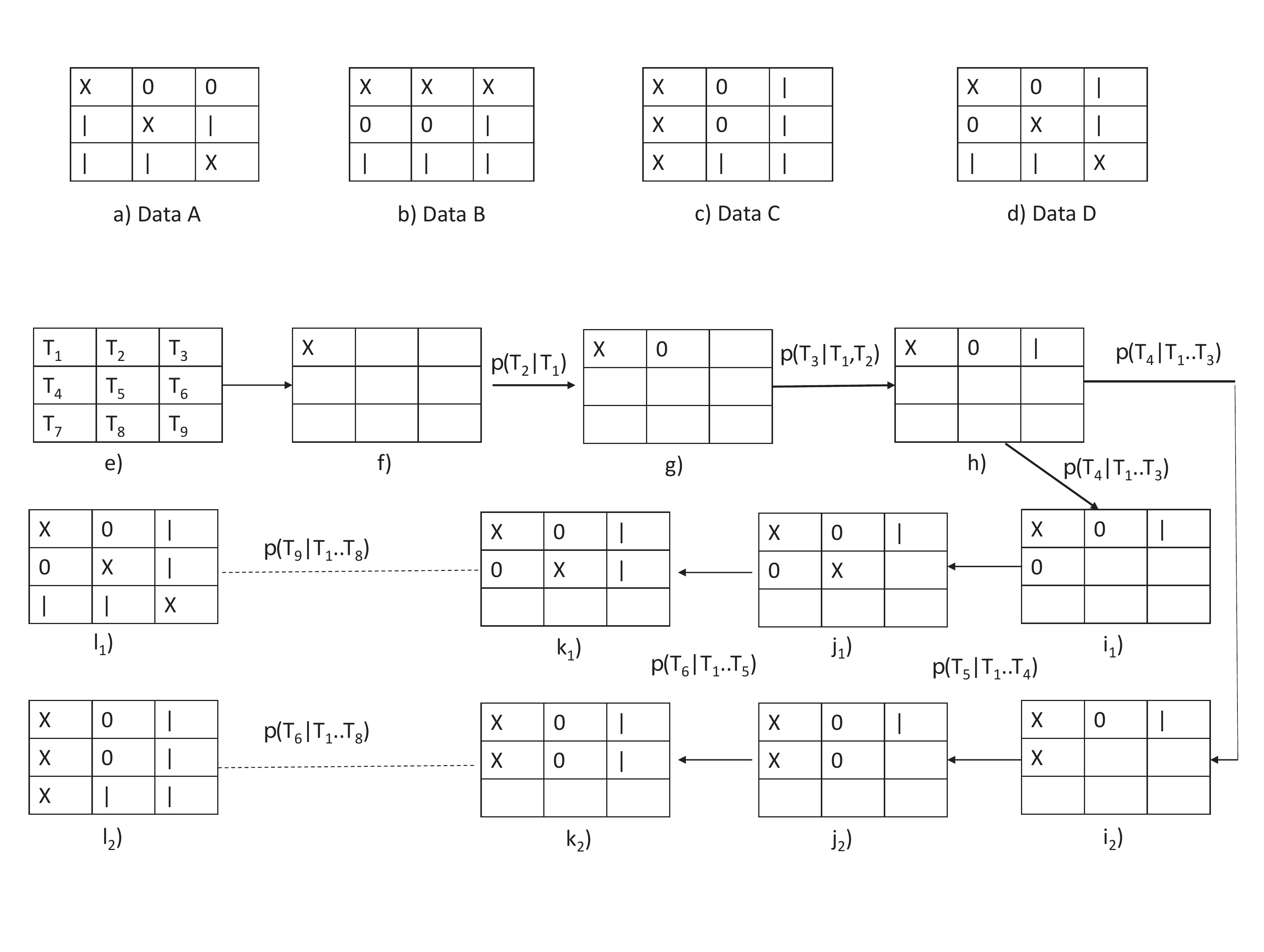}
	\caption{Example dataset with 4 elements and inferencing the dataset }
	\label{fig:exam}
\end{figure}

\subsection{Generative Modeling: an example}
Consider a data set with 4 elements as shown in Figs. (\ref{fig:exam}a - \ref{fig:exam}d).  Each element of this data set is a $3 \times 3 $ matrix whose entries are ordered from $T_1$ to $T_9$, where each of the element of the matrix comes from the set $S = \{ 0, \, \times, \, | \}$. Assume that we can estimate the joint probability distribution $p \left(T_1, T_2, \dots, T_9 \right)$ for the data in Fig. \ref{fig:exam}(a-d).  The inference after training is shown in Figs. \ref{fig:exam}(e- l). Suppose we observe (Fig. \ref{fig:exam}f) $T_1=\times$. We can now predict the values from $T_2, T_3$ to $T_9$ onwards conditional on $T_1=\times$ using the trained generative model.  Note that there are two paths to take once we observe $T_1=\times$, with different conditional probabilities. It can be easily inferred from the data (Figs. (\ref{fig:exam}a - \ref{fig:exam}d)), that probability that  $T_2$ will be $0$ is higher as compared from $\times$. Further on observing $T_2$ (which may not be same as what was predicted before), we update the probabilities for $T_3$ to $T_9$ and the sequential prediction process continues as we observe more variable. The next question therefore is, how to model the joint and the conditional distributions?
\subsection{Convolutional Neural Networks (CNN) for ETA}
Convolutional neural networks (CNN) is a class of deep learning models that has provided state of the art performance in various image/video classification tasks.  CNN captures the local spatial coherence by ``convolving" a local 2-D area with filters and thereby absorbing the spatial dependencies in an image. Intuitively, filters perform the task of feature extraction from the matrix (or an image).  Many filters can be passed through these matrices, each picking a different set of features. For example, a horizontal line or a vertical line filter.  Convolutional networks employ these filters, slices of the matrix feature space, and map them one by one. In other words, they create a map of each place where these features occur.  A general CNN architecture ( Figs. ~\ref{fig:m44}) has many layers. Following convolution operation through filters, the resulting matrix is passed through layers containing nonlinear transforms like $tanh$ or a rectified linear unit ($ReLU$) that is generally applied to each element of the matrix. \par
Similar to an image, one of the major motivation for using a CNN for the ETA estimation problem is the natural spatial and temporal correlation available in the ETA data of bus trips for a given route. However, using the ``regular" convolution filter in mask-CNN may imply that we end up using points for the convolution operation that may not have been generated yet. This implies that we may break the causality of the system.  For instance, to predict the ETA between stops $j-1$ to $j$, we cannot use the data for stops $j$ to $j+1$, as that trip has not happened yet. This challenge can be overcome by using an appropriate mask along with convolution operation that maintains the causality of the system.  Also, masking gives the flexibility to restrict the dependencies.  One of the contributions of our work is to automate the selection of an appropriate filter that decides the optimal dependencies for the ETA estimation task.
The dependencies that are captured using the CNN are nothing but the conditional probability distribution function that  we seek to obtain. We now discuss the proposed mask-CNN model and the ETA estimation problem in more detail.

%
\subsection{Mask-CNN architecture}
The mask-CNN architecture is a fully convolutional network of seven layers that preserves the spatial resolution of its input throughout the layers and outputs a conditional distribution at each location.
We first define the input provided for the training of the architecture. The time taken to travel between any two consecutive stops in the evening may not be dependant on the time taken in the morning. Therefore, we can divide everyday trip data into smaller overlapping chunks of window size $H$. 

Let $K$ denote the  total number of stops in a route.  Similar to an image, we define the collected bus ETA data as a 2-D matrix of dimension $H \times K$, one for each day, whose rows are the trips on a route in a day and the columns contain the travel time between two consecutive stops.  In the case of an image, a pixel generally takes value in the range of $0$ to $255$. The traffic ETA matrix can be seen as an image with ETA values ranging from $0$ to $C$ (seconds). The value of $C$ is decided based on the maximum possible value of ETA and quantization levels $l$.

\begin{figure}[ht]
	\centering
	\includegraphics[width=1\linewidth]{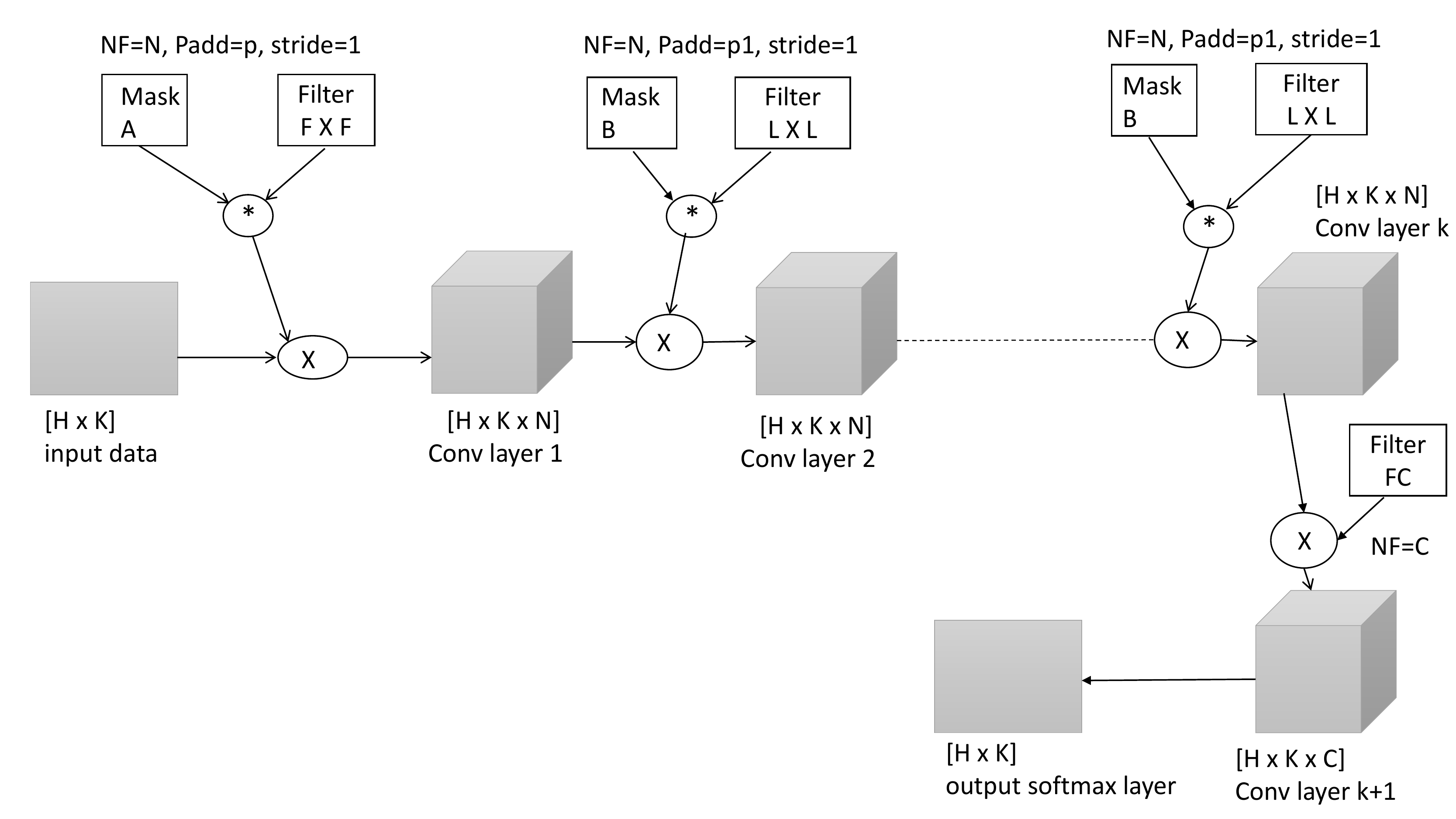}
	\caption{Architecture of mask-CNN}
	\label{fig:m44}
\end{figure}

The architecture of mask-CNN is shown in Fig. \ref{fig:m44} where the input to the model is $H \times K$ ETA matrix and the corresponding output is a $H \times K \times C$ tensor. Here $H$ is the window size for the number of routes and $K$ is the number of bus stops. Applying a softmax layer on the above tensor generates an output tensor  $H \times K \times C$ corresponding to the probabilities of each pixel taking $C$ values. Finally, the value with maximum probability is chosen.  The first layer is a mask A CNN layer with filter dimension $F\times F$ with a total of $N$ filters, padding as $p$ and stride as 1.  See Fig. \ref{fig:mask}. Next $k - 1 $ layers after the first layer is Mask B layer with filter dimension $L\times L$ with $n$ number of filters, padding as $p_1$ and stride as 1. ReLU activation function is used after every convolution layer. The last convolution layer $FC$ is a fully connected layer with filter size 1.  The number of filters in the fully connected layer $FC$ is equal to $C$. The end layer in the mask-CNN layer is the softmax layer which assigns the probability to all the discrete variables $C$ and output is the discrete variable with the highest probability.

The overall architecture of the masked CNN for traffic state prediction is as follows
\begin{enumerate}
	\item First layer is the Mask A CNN layer with filter dimension $F \times F$. 
	\item There are $k-1$ Mask B CNN layers with filter dimension $L \times L$. 
	\item $ReLU$ activation is followed by every convolution layer.
	\item At the output stage there is a fully connected convolution layer followed by a $C$-way softmax layer. 
\end{enumerate}
\subsubsection{Masking}

\begin{figure}[ht]
	\centering
	\includegraphics[width=1\linewidth]{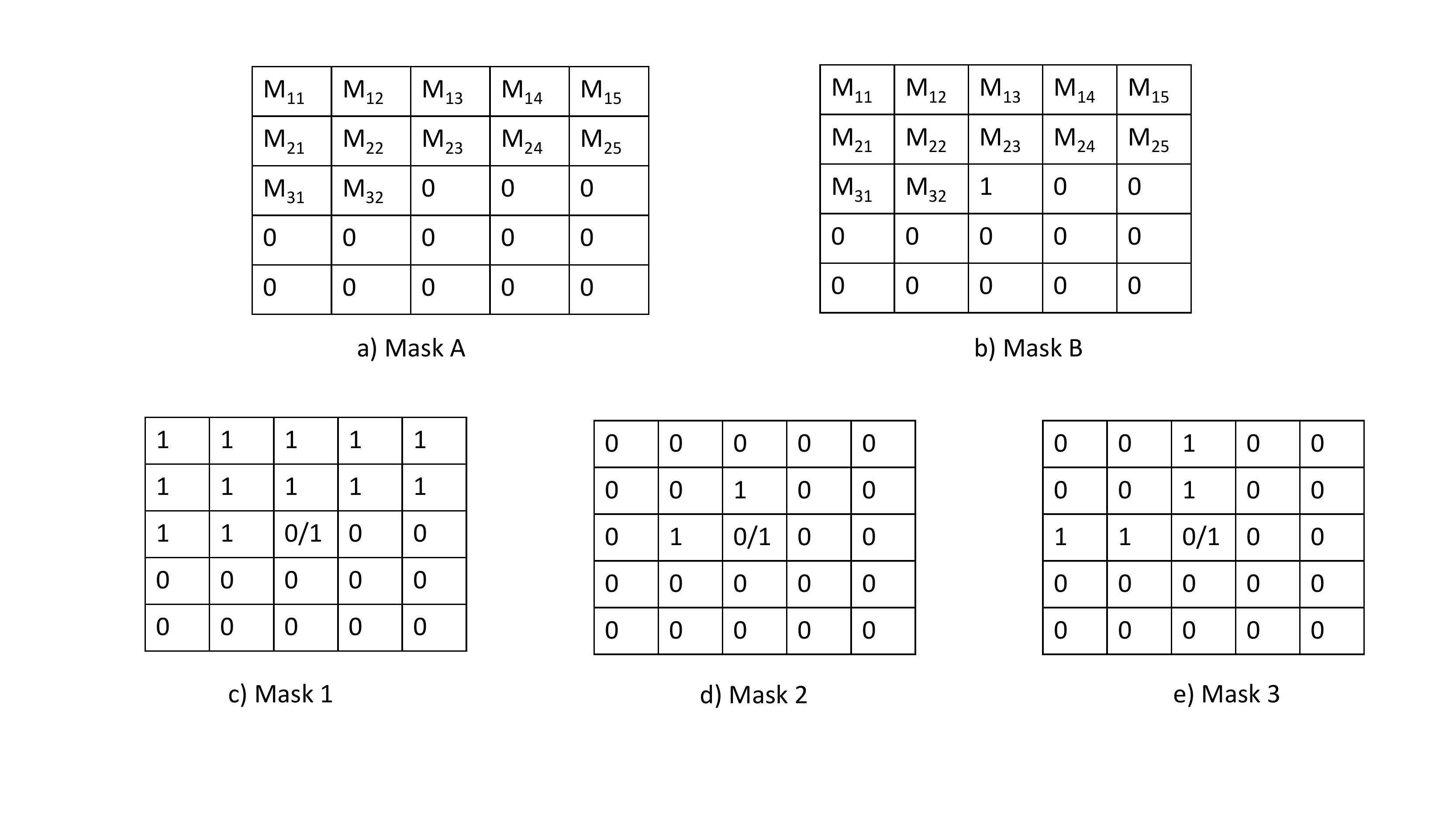}
	\caption{Different masks used in mask-CNN}
	\label{fig:mask}
\end{figure}

There are two mask used in the masked CNN, mask A and mask B. Mask A is the first layer in the mask CNN and shows the effect of already predicted ETA points on the point that we are about to predict sequentially. Mask B is used in rest of the layers. For mask B the connection with the about to be predicted pixel point is also included. Mask A and B for $5 \times 5$ filter are shown in Fig. \ref{fig:mask} (a, b) where $0$ denotes that the future dependencies are removed from the prediction.  In mask A and mask B, the entries $M_{i,j}$ can be 0 or 1 based on how we want to model the past dependencies for the next prediction. We employ three Masks (Mask 1, Mask 2 and Mask 3) for ETA prediction with different $M_{i,j}$. 
\subsection{ETA prediction using the trained model}

\begin{figure}[ht]
	\centering
	\includegraphics[width=1\linewidth]{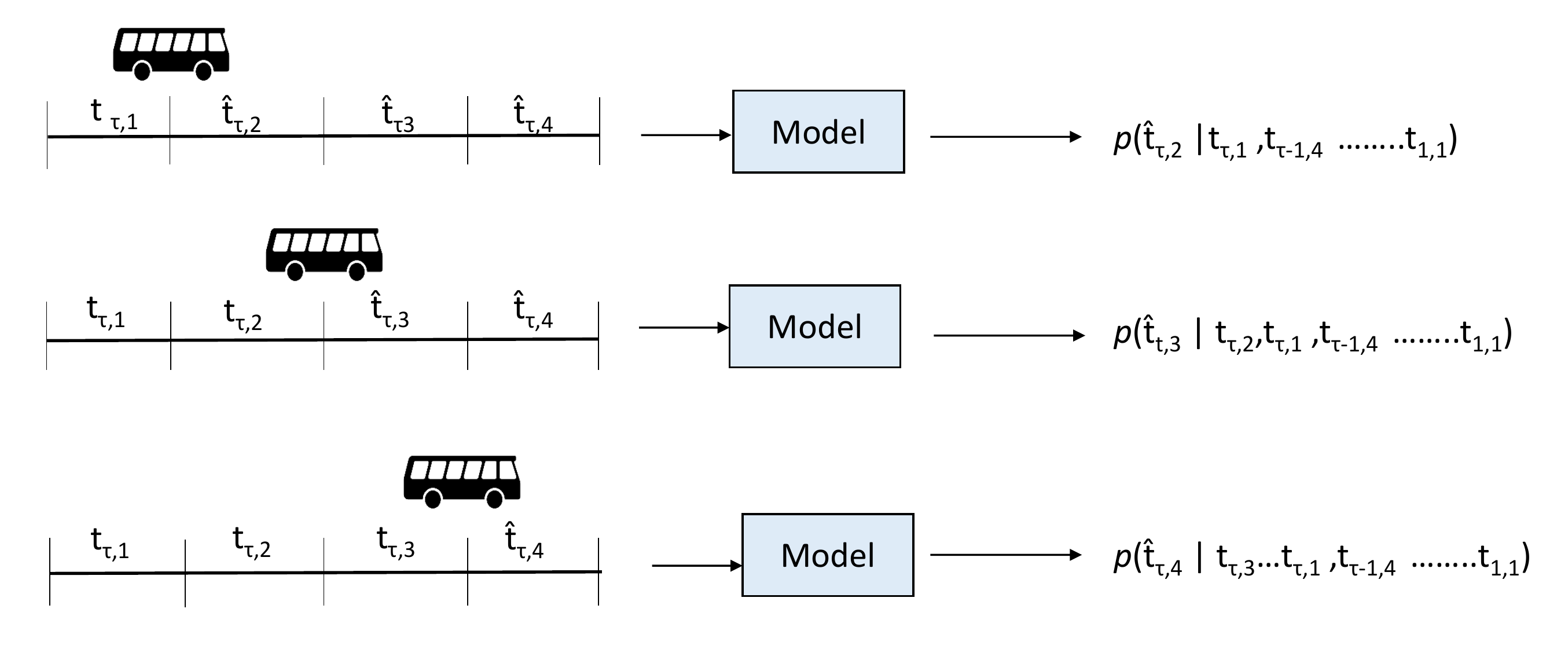}
	\caption{Inferencing the Travel Time}
	\label{fig:m3}
\end{figure}

Once the model is trained using the historic travel time data, we are now ready to provide ETA estimation for every trip in the route.  Fig. \ref{fig:m3} explains the inference process with a simple example of a route with 4 stops for the $t^{th}$ trip. At the beginning of the trip, the ETAs for various stops is generated using sampling from the joint distribution that is trained using historic data. However, as soon the bus crosses the first stop, the subsequent sequential generation of ETAs would take into account the actual travel time $t_{t,1}$ to the first step. Similarly, the ETAs are updated when the trip crosses second and the third stops. We predict the ETA as:
\begin{equation}
    \hat{t}_{tr,(k+1)}\leftarrow \textbf{Model}\,\left( {t}_{tr,(1:k)}  \,\,,{t}_{(1:tr-1),(1:K)}\right)
\end{equation}
\section{Results}
\begin{figure}[ht]
    \centering
    \includegraphics[width=1\linewidth]{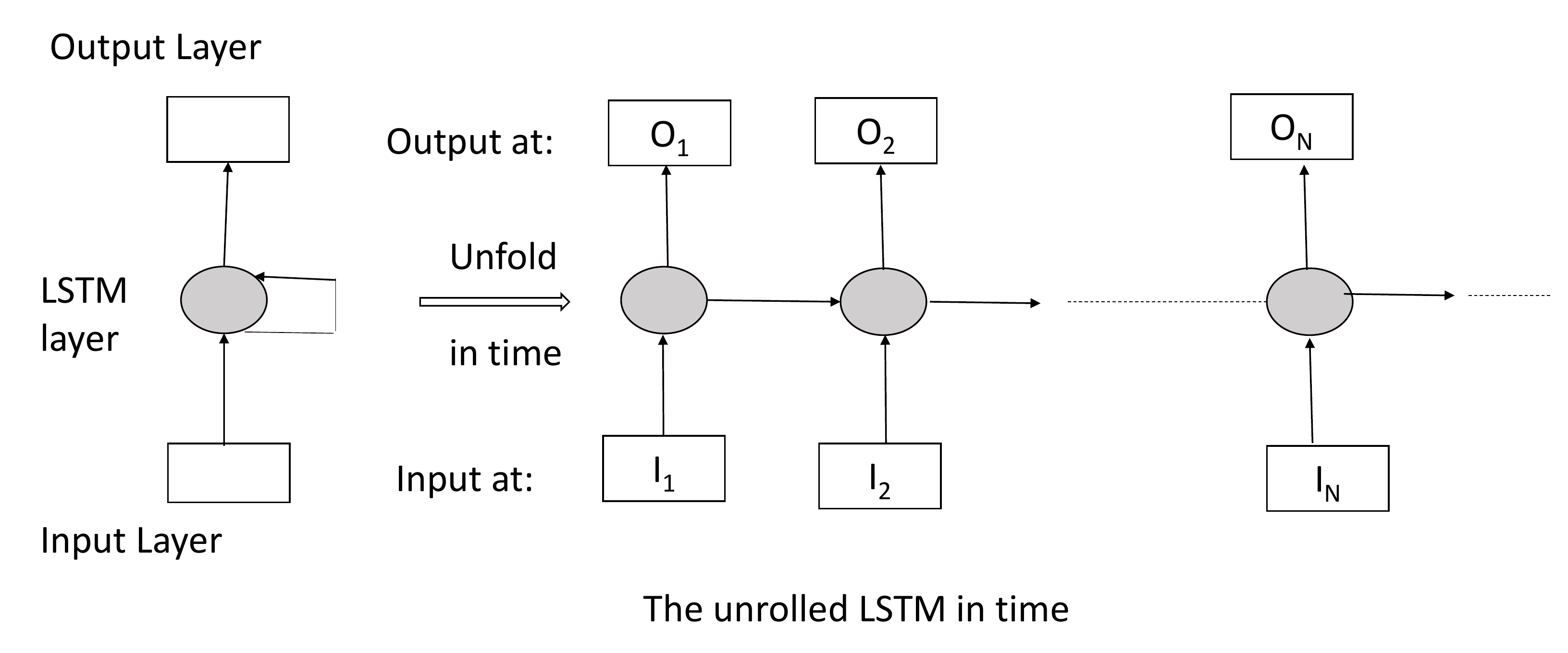}
    \caption{The LSTM architecture is unrolled along time to describe a complete trip}
    \label{fig:lstm}
\end{figure}

We now discuss the performance of the proposed mask-CNN algorithm for the ETA estimation task for a bus route network. We compare our technique with the state-of-the-art approaches like time series prediction, deep learning, as well as the matrix completion approaches below:
\begin{enumerate}
    \item ARIMA (Autoregressive Integrated Moving Average) \cite{billings2006application}.
    \item LSTM (Long Short Term Memory) \cite{duan2016travel} is one of the recent methods to compute the ETA in public transit as well as cabs. We used the architecture shown in fig \ref{fig:lstm}. 
    \item VBSF (Variational Bayesian Subspace Filtering) \cite{vbsf}: Online matrix completion frameworks are not only employed to fill the missing entries of a matrix but also for prediction of the future columns. VBSF is one of the matrix completion algorithm that is used for traffic estimation and prediction and is shown to outperform other similar techniques. 
\end{enumerate}

\subsection{Dataset}
\begin{figure}[ht!]
	\centering
	\includegraphics[width=0.8\linewidth]{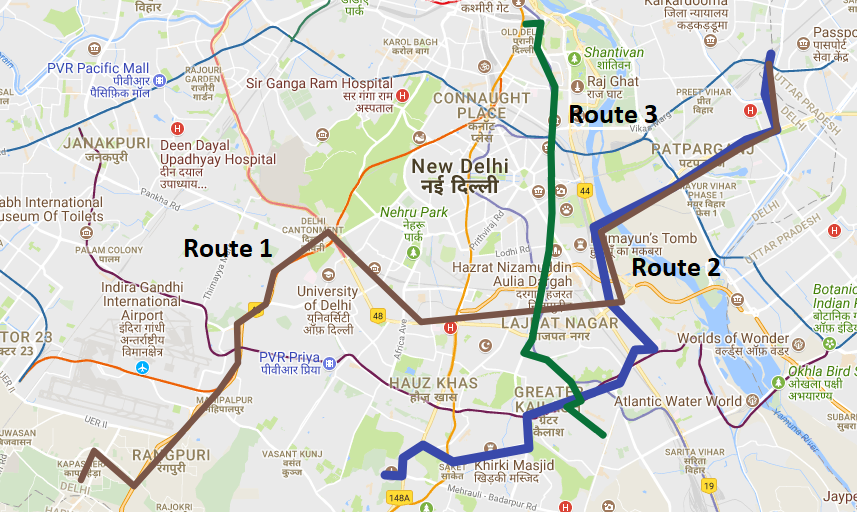}
	\caption{Routes used for data collection}
	\label{fig:map}
\end{figure}

Our dataset consists of travel time information of three bus routes in New Delhi as drawn in Fig. \ref{fig:map}. The lengths of these routes are approximately 30 km, 22 km and 20 km. Each route operates around 40 trips per day with nearly fixed starting timetable. The bus route is made of a sequence of stops, and we collect the arrival time and the departure time for these stops. We divide everyday day into smaller overlapping chunks of $h=10$ trips. Of the three months of collected data, we use two months of data for training and the third month data is used for evaluation of all the algorithms. 
\subsection{Training Parameters}

We employ a variety of masks based on the dependencies we want to capture in the dataset. We use three different kinds of the mask in our evaluation (mask A1 and B1 for mask 1, mask A2 and B2 for mask 2, mask A3 and B3 for mask 3).
The masks 1, 2 and 3 for filter dimension 5 is shown in Fig. \ref{fig:mask} (c, d, e) respectively, where the middle element in Fig. \ref{fig:mask} (c, d, e) is 0 for mask A and 1 for mask B. \par
To train our model, we use a batch size of 32 with a learning rate of 0.01 along with RMSprop optimizer \cite{ruder2016overview}.  We test three filter dimension values of 7, 5 and 3 for both $F$ and $L$  in mask A and B. Also, the number of blocks $k-1$ for mask B is set to 6. The number of classes for softmax layer $C$ is taken as 128, 256 and 512. Stride in the CNN is taken as 1, while zero padding ($p$, $p1$) for filter size 7 is taken as 3, for filter size 5 is taken as 2, and for filter size 3 is taken as 1. We first remove the outliers from the training data and fix the maximum ETA for a stop as 1024 for our data. The number of classes $C= 1024/l$ is decided based on the travel time data where $l$ is the quantization level . In our case, we fix the value of $C$ using the grid search as shown in Table \ref{tab1:tune}.
\subsection{ETA Estimation}
We use the standard mean absolute percentage error (MAPE), root mean squared error (RMSE) and mean absolute error (MAE) as our performance metrics defined as follows:
\begin{equation*}
      \text{MAE}= \frac{1}{T\,K}\sum_{k=1}^K\sum_{\tau=1}^T {| \hat{t}_{\tau,k}-t_{\tau,k}|}
\end{equation*}
\begin{equation*}
      \text{MAPE}= \frac{1}{T\,K}\sum_{k=1}^K\sum_{\tau=1}^T \frac {| \hat{t}_{\tau,k}-t_{\tau,k}|}{| t_{\tau,k}|} \times 100\%
\end{equation*}
\begin{equation*}
      \text{RMSE}= \frac{1}{T}\sum_{\tau=1}^T\sqrt{\frac{\sum_{k=1}^K {( \hat{t}_{\tau,k}-t_{\tau,k})}^2}{K}}
\end{equation*}

The comparative performance of different filters, masks, and number of classes are shown in Table \ref{tab1:tune}.
Based on these results, we tune the filter dimension as 5, mask as 2 and softmax classes as 512.. Mask 2 and filter dimension of 5 performs better than the other mask and filter because there are lesser dependencies of the road segments far away from the predicted road segment. Note all these parameters can easily be auto-tuned to make the system manual-tuning free. \par
\begin{table}[ht!]
	
	\begin{center}
		\begin{tabular}{ll}
			\hline 
			
			&MAPE\\
			\hline
			Filter 3 ,mask 1 , classes-256 &0.2927 \\
			Filter 3 ,mask 2 , classes-256 &0.2815 \\
			Filter 3 ,mask 2 , classes-512 &0.2619 \\
			Filter 5 ,mask 1, classes-512 &0.27114 \\
			Filter 5 ,mask 2  , classes-512 &0.23991 \\
			Filter 5 ,mask 3 , classes-512 &0.24208 \\
			Filter 7 ,mask 2 , classes-512 &0.27078 \\\hline
			
		\end{tabular}

	\end{center}
	\caption{Performance comparison for different filter size and Quantization classes}
	\label{tab1:tune}
\end{table}

\begin{figure}[ht!]
	\centering
	\includegraphics[width=1\linewidth]{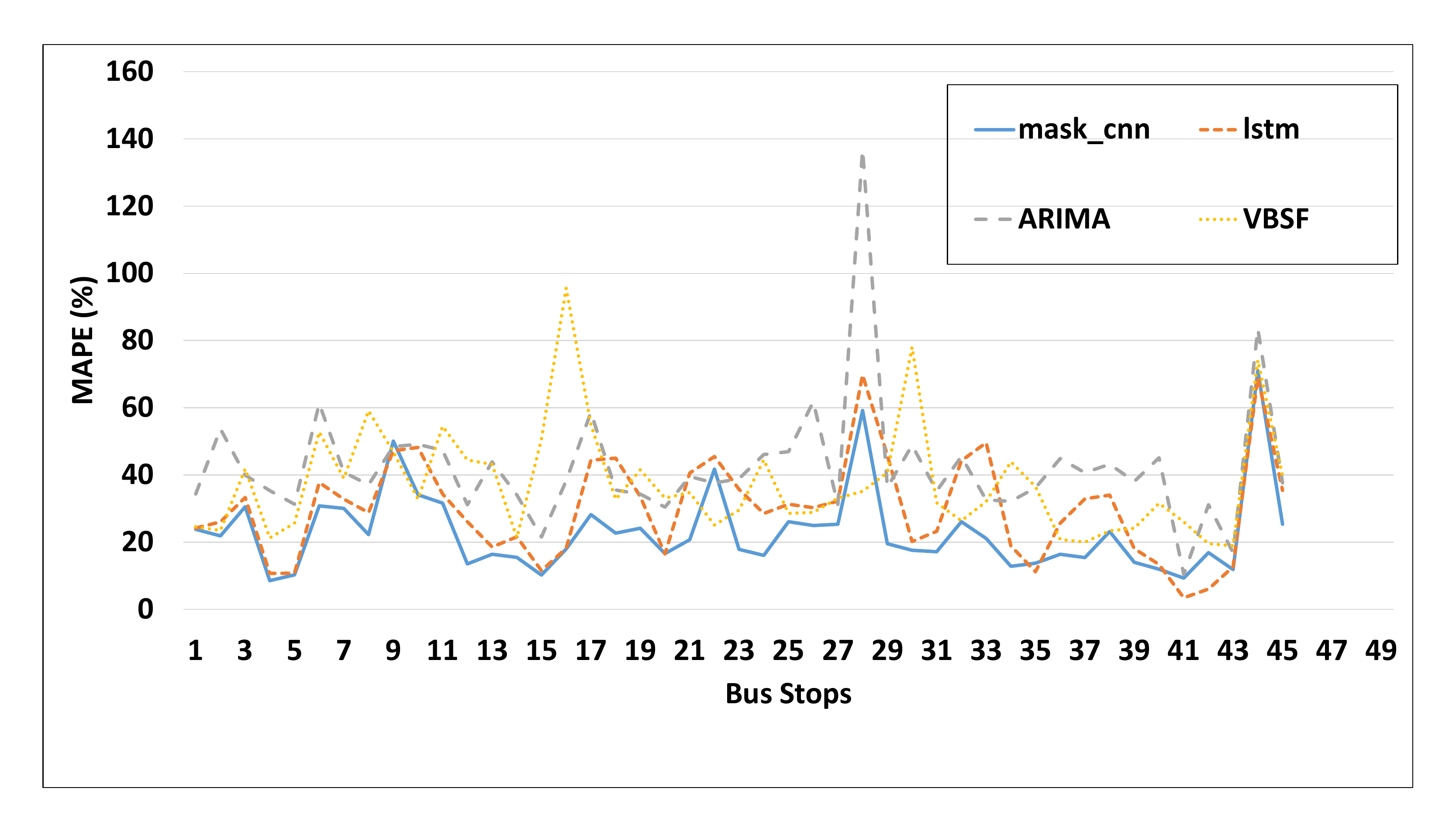}
	\caption{Comparison of Masked CNN for a bus route}
	\label{fig:res3}
\end{figure}
To evaluate the performance of the mask-CNN approach, we compare it with ARIMA, LSTM and VBSF, since these methods outpeform other methods in their class. Comparing our approach with other deep learning approaches is not possible due to a relatively smaller data set that we deal with. Further as mentioned before, we do not possess other parameters like driver id, weather information, etc. that methods like \cite{wang2018will, deepTTE} use. Finally, none of the methods use the online prediction mechanism as the trip progresses. 
Table \ref{tab:comp} and Fig. \ref{fig:res3} provide the required comparison.  Fig. \ref{fig:res3} shows the variation in performance concerning all the stops in the route for the first route. The other two routes behave similarly.  LSTM is popular method used for the time series data. We show that mask-CNN performs better than LSTM for the bus ETA problem. For eg. in case of route 1, the average ETA for the stops are 150 secs. There are 7 stops with standard deviation higher than 100 secs and 16 stops with  standard deviation less than 50 secs. The prediction for the stops with high variation contribute to high error. On an average the error in the ETA prediction for LSTM is 38 secs while for mask-CNN is 30 sec. We demonstrate that our algorithm performs better than the ARIMA, LSTM and VBSF in most of the cases. \par
The advantage of the mask-CNN model is that it is not designed empirically for different times. Our model does not require any information regarding the time, day, driver profiles and other complex features required by other models \cite{wang2018will, deepTTE} to model the ETA prediction. The only information to train the model is the travel time data between stops. Mask-CNN captures the dependency in the data temporally as well as spatial by representing the ETA as a image and modeling the same with different masks. 

\begin{table}[!ht]

	\begin{center}
		\begin{tabular}{l l l l l l l}
			\hline
			& &Route 1&& Route 2&& \\
			&MAPE &MAE &RMSE &MAPE &MAE &RMSE \\ 
			&(\%)&(sec)&(sec)	&(\%)&(sec)&(sec)	\\\hline
			ARIMA&48.64&53.77&69.42&68.42&73.84&84.18\\ \hline
		
			VBSF&37.02&50.62&65.27&74.96&99.35&121.26\\ \hline
			LSTM&29.587&38.59&56.65&52.82&65.76&79.97\\ \hline
			mask-CNN&23.991&30.40&45.84&46.24&62.15&76.60\\ \hline

		\end{tabular}
	\end{center}
	\begin{center}
		\begin{tabular}{l l l l }
			\hline
			& &Route 3&\\
			&MAPE &MAE &RMSE  \\ 
			&(\%)&(sec)&(sec)	\\\hline
			ARIMA&47.44&49.37&60.75\\ \hline
			VBSF&45.2&48.31&59.05\\ \hline
			
			LSTM&41.09&43.25&52.15\\ \hline
			
			mask-CNN&36.04.991&39.42&49.89\\ \hline
		\end{tabular}
	\end{center}
		\caption{Performance Comparison}\label{tab:comp}
\end{table}
\section{Conclusion}
In this paper, we investigate a deep learning based generative model to estimate the ETA of a bus trip in real time.  We train a model for each individual route using historical data of trips collected over two months.  We observe that we could learn a reasonably accurate joint distribution of the ETA variables across day and bus stops. Our model is easy to implement for transit agencies, adaptive and utilises the real-time information of the trip as well.  It has a great potential to be used in places where other dense traffic data set is not available. For future work, we planned to explore the model for unexpected events. Also, we are interested in determining the uncertainties in our prediction results.  
\bibliographystyle{IEEEtran}
\bibliography{p}
\end{document}